\documentclass[conference]{IEEEtran}

\usepackage{cite}
\usepackage{amsmath,amssymb,amsfonts}
\usepackage{algorithmic}
\usepackage{graphicx}
\usepackage{textcomp}
\usepackage{xcolor}
\usepackage{url}
\usepackage{hyperref} 
\usepackage{fontawesome}
\usepackage{comment}
\usepackage{caption}
\usepackage{subfig}

\begin{document}

\title{Robust Sensor Fusion Algorithms Against Voice Command Attacks in Autonomous Vehicles}

\author{\IEEEauthorblockN{
Jiwei Guan\IEEEauthorrefmark{1},
Xi Zheng\IEEEauthorrefmark{1},
Chen Wang\IEEEauthorrefmark{2}, 
Yipeng Zhou\IEEEauthorrefmark{1} and
Alireza Jolfaei\IEEEauthorrefmark{1}}
\IEEEauthorblockA{\IEEEauthorrefmark{1}Department of Computing, Macquarie University, Sydney, Australia
\\jiwei.guan@hdr.mq.edu.au, \{james.zheng, yipeng.zhou, alireza.jolfaei\}@mq.edu.au}
\IEEEauthorblockA{\IEEEauthorrefmark{2}CSIRO Data61, Sydney, Australia
\\chen.wang@data61.csiro.au}}


\maketitle
\begin{abstract}
With recent advances in autonomous driving, voice control systems have become increasingly adopted as human-vehicle interaction methods. This technology enables drivers to use voice commands to control the vehicle and will be soon available in Advanced Driver Assistance Systems (ADAS). Prior work has shown that Siri, Alexa and Cortana, are highly vulnerable to inaudible command attacks. This could be extended to ADAS in real-world applications and such an inaudible command threat is difficult to detect due to microphone nonlinearities. In this paper, we aim to develop a more practical solution by using camera views to defend against inaudible command attacks where ADAS are capable of detecting their environment via multi-sensors. To this end, we propose a novel multimodal deep learning classification system to defend against inaudible command attacks. Our experimental results confirm the feasibility of the proposed defense methods and the best classification accuracy reaches 89.2\%. Code is available at \url{https://github.com/ITSEG-MQ/Sensor-Fusion-Against-Voice-Command-Attacks}.
\end{abstract}

\IEEEpeerreviewmaketitle

\section{Introduction}
With the aid of machine learning, Advanced Driver Assistance Systems (ADAS)\cite{bengler2014three} has acceptance and recently emerged. In ADAS, Voice Control Systems (VCS) are currently being explored both in industry and academia as such systems can provide human-vehicle interaction in a faster and more convenient way \cite{baidu2020apollo,google2020waymo}.
A VCS enables drivers to communicate with their phones or computers in vehicles so that they can get directions, send emails, make phone calls, or play music through voice instructions. With projects such as Baidu Apollo \cite{baidu2020apollo} and Google Waymo \cite{google2020waymo} continuously reshaping voice controllable functions in vehicles, the VCS technology is able to control embedded systems \cite{tdax2015embed} and engine control units, and even communicate with other vehicles in the near future. Thus, the potential benefits of voice control will be much greater than just finding restaurants and searching for directions, particularly for Level 5 vehicles, which are fully automated and operate on the road with no human interactions.

\subsection{ADAS Attacks}
The primary goal of ADAS is to ensure safety and reliability for drivers and vehicles. However, recent examples of ADAS failures include accidents in Uber \cite{uber2020accident} and Tesla \cite{tesla2020accident} which raise serious security concerns \cite{petit2015remote} in autonomous driving. Such security threats to autonomous driving take the form of malicious interference with navigation, steering, and control in real-life traffic conditions. 
In the quest for continuous improvement, ADAS technical developments have become quite mature and more complex.
Such rapid growth requires significant efforts and presents a major challenge for ADAS defence. 
There are many research works on the attacks of real ADAS systems, including object detection systems, LiDAR systems, global navigation satellite systems, semantic segmentation systems, speech recognition systems and speaker recognition systems etc. For example, attackers can spoof the sensor networks on the vehicle, modify the internal communication channels, accessing the cryptosystems without legal permission, steal unprotected ADAS data, manipulate firmware that has no security updates until the system fails, and conduct audio adversarial attacks \cite{yan2016can}. In this paper, we  mainly focus on voice command attacks which are difficult to detect.

\subsection{VCS Attacks in ADAS}
Since ADAS are vulnerable to attacks resulting in dangerous misbehavior, it is of great importance to detect ADAS attacks quickly with high accuracy. However, such aim is difficult to achieve. For instance, recent works \cite{zhang2017dolphinattack}have shown that hidden and inaudible voice attacks are feasible and such attacks can fool commercial voice recognition systems such as Apple Siri and Google Assistant. There are no existing real-time viable solutions to detect such attacks. On the other hand, multi-modality sensor fusion algorithms have been widely adopted for Unmanned Aerial Vehicle (UAV) landing and autonomous driving needs to provide requested robustness and safety assurance \cite{feng2020deep}. Based on this finding, we utilize multiple state-of-the-art sensor fusion algorithms to leverage voice command with the existing in-vehicle camera. We would like to explore whether such multi-modality module for voice command can be robust enough to defend against inaudible voice attacks, 


More specifically, we make the following three main contributions:
\begin{itemize}
\item \emph{Attack Model}. We present a highly likely attack scenario where inaudible voice attacks can be possible for autonomous vehicles. 

\item \emph{Multi-Modality Defence Models}. We design a few multi-modality neural networks taking both voice commands and camera input images and outputting the filtered voice command (removing the inaudible voice attack commands).

\item \emph{Experiments}. We conduct a few experiments based on real-world traffic sign and speech command datasets to evaluate our proposed multi-modality defence models. Under attacks, the best performing model can still reach satisfying results with 89.2\% for both accuracy and F1. Without attacks, the best performing model can reach 99\% for both accuracy and F1, reducing inaudible voice attack success rate to roughly 10\%. 

\end{itemize}

The remainder of the paper is organized as follows. Section~\ref{sec:related_work} presents current audio physical attack and its related defense. Section~\ref{sec:threat_model} introduces the threat assumption and motivating scenario. Section~\ref{sec:methdology} describes our proposed defense architecture. Section~\ref{sec:experiment} explains the dataset, experiment and evaluation results. Section~\ref{sec:discussion} discusses the advantage and limitations of a series  of experiments. Finally, Section~\ref{sec:conclusion} summarizes our paper.


\section{Related Works}
\label{sec:related_work}
In this part, we review the existing literature on
audio physical attacks and the corresponding defence methods.

\subsection{Audio Physical Attacks}
Recent works have shown that ultrasound is designed as carrier for hidden and inaudible command attacks \cite{song2017poster,zhang2017dolphinattack,roy2018inaudible}. These attacks have demonstrated that the higher frequency sounds are inaudible to human hearing and undetectable to VCS. They are sufficiently sophisticated to fool commercial speech recognition products such as Apple Siri, Google Assistant, and Amazon Alexa if the endpoint is using Micro-Electro-Mechanical System (MEMS) microphone. For example, a voice command like ``Hi Siri, start my car'' in Car Play would be a serious threat, which is inaudible.
More specifically, 
Song \textit{et al.} \cite{song2017poster} proposed a novel attack approach through ultrasound to control VCS, relying on the non-linearity of microphone circuits. The modulated function signal could demodulate to the lower frequency normal voice that was received by VCS then converted to text. The Dolphin Attack \cite{zhang2017dolphinattack} launched voice commands using ultrasonic carriers to inject VCS across a wide range of targeted devices, including Audi vehicles. Subsequently, Roy \textit{et al.} \cite{roy2018inaudible} extended this idea by using a custom-designed speaker to expand the attack range in a large room or a house with open windows. They employed sound pressure levels as a defense approach to classify the attack. The overall accuracy was 99\% in all experiments although they noticed leakage audibility and reported failed defenses. The SurfingAttack \cite{yan2020surfingattack} showed that ultrasonic waves leveraged in the context of solid media achieved multiple rounds of inaudible communication with VCS across a long distance. Sugawara \textit{et al.} \cite{sugawara2020light} manipulated laser signal injection into the targeted commercial VCS at a long distance over 100 meters. It also proved the MEMS microphone was vulnerable to light malicious commands. In another work, CommanderSong \cite{yuan2020commandsong} tried a novel technique to hide voice commands into songs when playing and successfully injected VCS without being detected. In \cite{zhang2019using}, it constructed a clone of normal dialogue to craft the activation word in a way that would not be detected.

Inaudible voice commands are both the targeted and black box attack but requires playback records or text-to-speech to synthesize new audios. It cannot use and modify existing pre-recorded audio in real-time. More importantly, all above mentioned empirical studies show that malicious voice command attacks are relevant to hardware without any algorithms, which does not always require machine learning systems.

\subsection{Audio Adversarial Defense}
To defend against these audio attacks, there are some existing defences.
In \cite{carlini2016hidden}, authors used a Levenshtein edit distance to measure phoneme sequences of two transcriptions as detection techniques. However, they found that the adversary could mask these defense mechanisms. Mendes and Hogan \cite{mendes2020defending} reverted an adversarial audio example at each frequency value back to its original transcription by adding Gaussian noise. Active noise cancellation was conducted to defend inaudible voice command attacks \cite{he2019canceling} and the method was to use a signal transmitter to create a special sound spectrum that neutralizes the attack signal via software design. This method broadcast inaudible ultrasonic signals as positive and normal signals as negative, then used a short time Fourier transform to determine the attack frequency. Moreover, the work \cite{zhang2017dolphinattack} reported that the attack signal was the difference between the real signal and the Google text-to-speech signal. It proposed a machine learning algorithm: Supported Vector Machine (SVM) to find unique features of modulated voice commands. Authors managed 24 samples: 5 of normal audio as positive samples and 5 of recovered audio as negative samples, with the other 14 being the test dataset. The prediction reached 100\% true positive rate and 100\% true negative rate. However, the whole dataset was not quite large enough to use in machine learning. In light commands attack \cite{sugawara2020light}, they suggested using an additional layer of authentication, device locations to prevent eavesdropping and use a multi-microphone array for sensor fusion due to the laser's inability to attack a microphone array simultaneously. 

Many researchers have devised security solutions for defending against audio adversarial attacks. In this part, we discuss the hardware perspective. First, a MEMS microphone is the basic hardware that can receive ultrasonic sound over 20kHz. Most smart devices use an embedded MEMS microphone. Therefore, the ideal design for inaudible command attacks is to deny specific higher frequency sounds. Enhancing the MEMS microphone for suppression and detection in the ultrasound range is a solution to this form of attack. Second, the microphone hardware adds functional module in MEMS that can filter out higher-frequency sound. Given legal and illegal voice commands, the module can detect the signals within the ultrasound frequency range that exhibits amplitude modulation, then demodulate the signals to obtain the baseband \cite{zhang2017dolphinattack}. This process is known as inaudible voice command cancellation. However, both of the above defense strategies are difficult to implement in commercial hardware products. 


Unfortunately, very few effective defense mechanisms are available for inaudible command attacks today. None of the above defense mechanisms sufficiently employ comprehensive visual semantic information for action prediction. It is worth highlighting that much work has been done to defend against audio adversarial attacks, but none of the studies have involved multi-modality methods. The motivation behind the defense mechanism operated by multiple modalities is to determine whether feature fusion is a practical countermeasure.

\section {Threat model}
\label{sec:threat_model}
As mentioned previously, ultrasound is a type of high-frequency acoustic wave that is beyond human hearing (above 20-kHz), but smart devices can receive the inaudible signal by MEMS microphone. Such a command attack is designed to inject MEMS microphone sensors using ultrasound as a carrier to ensure inaudibility. Thus, this type of attack can be used in a malicious application without notification and authorization. Unfortunately, the VCS in ADAS using such MEMS microphone can receive inaudible speech commands.

\subsection{Attack Assumption}
The attack assumption is that emitting inaudible physical signals could spoof the microphone in the VCS. A number of studies have investigated such audio adversarial attacks scenarios. Specifically, The inaudible attack is easily received by VCS and can control ADAS without any sensors being damaged or the system being in any way physically altered. These attacks have the potential to cause significant damage and are difficult to detect because they are stealthy with the adversary likely to be mobile, contactless and at a distance from the target (vehicle). This paper extends the previous literature by focusing on developing a reliable technique against inaudible command attacks on ADAS.  

Here we assume a driver quite often interacts with a VCS when driving, and there is less interaction when the vehicle is parked. When the vehicle is driving on the road or in a quiet environment, acoustic transmissions can be actively received. Attacks are most likely to be successful in these environments. We also assume that the adversary has knowledge of the characteristics of the target VCS but not the complete knowledge that is typical of a white-box attack. Our scenario is consistent with a black-box attack, in which an adversary tries to use an inaudible voice command to manipulate the VCS. The adversary can target the VCS in an ADAS without direct access to the victim’s vehicle and cannot physically set or touch VCS settings. This procedure allows the adversary to communicate with the VCS while 
attacking the vehicle at an effective distance and gaining control of the ADAS without permission. Consequently, a Level 5 autonomous driving vehicle faces a high likelihood of being the victim when an inaudible command attack is conducted on its VCS.

\subsection{Attack Scenario}
\begin{figure}[htbp]
\centering
\includegraphics[width =.5\textwidth]{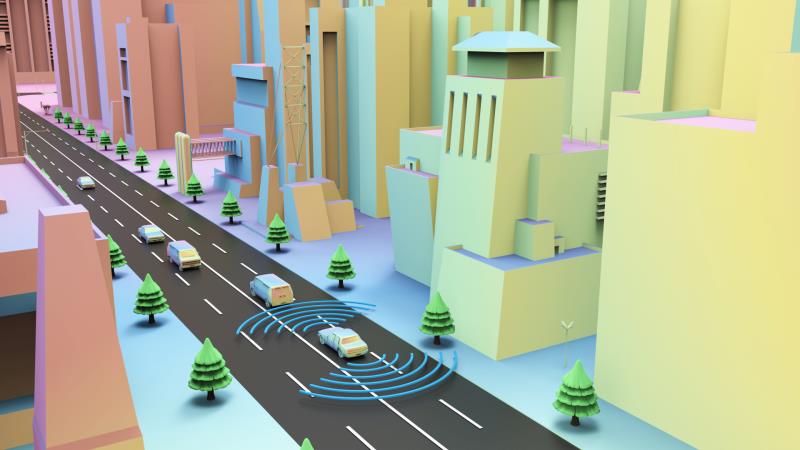}
\caption[Mobile audio attack]{Mobile attack scenario on a road.}
\label{fig 1} 
\end{figure}

The inaudible command attack for ADAS scenario on the road is shown in Fig. \ref{fig 1}. Let’s consider the sedan can send inaudible voice commands using prepared hardware to attack the van that is the victim. In addition, the attack firstly produces the inaudible activation command that meets the attack conditions, then creates an inaudible speech command that can be executed and correctly transcribed on VCS to ADAS to manipulate autonomous driving. Subsequent inaudible commands can be executed in silence. Once the victim van receives the inaudible commands, the adversary can take over the VCS on the ADAS, which can illegally access the automated driving application. In this threat model, the adversary’s goal is to unlock the VCS identification system without being detected. Because environmental noise can be easily heard by the public, the adversary plays pre-recorded audio by text-to speech (TTS) and sends the inaudible voice commands through an ultrasonic speaker \cite{zhang2017dolphinattack}. Through this method, the adversary can gain control of VCS at a lower cost.

Note that though the attacking vehicle is required to follow the victim’s vehicle, directly behind or in close proximity, such attack is dangerous and an effective system shall be built inside ADAS to sense such attack.  

\section{Methodology}
\label{sec:methdology}
\subsection{Multimodal Fusion}
Sensor fusion is the integration of multi-source data to avoid single data bias for reliable and accurate information. Fused data is the core and highly valuable information that improve robustness. Meng \textit{et al.} \cite{meng2020survey} summarized recent work of machine learning for data fusion. On the other hand, multi-modal perception data is complementary to each other, especially in autonomous driving scenario. Nabati and Qi proposed \cite{nabati2021centerfusion} center-based radar and camera fusion for detecting and tracking the surrounding objects in ADAS, which is a critical part of the perception system. Furthermore, information about a phenomenon can be acquired from a multimodal approach, in which various degrees of perceptions are provided. A previous study \cite{lahat2015multimodal} established that the rich characteristics of natural phenomena provide more complete knowledge than a single modality. That meant multimodal information could overcome the limitations of a single perceptual modality. 
Based on this observation, we use a multimodal model to defend against inaudible command attacks. Moreover, the defense strategy requires the fusion technology to be based on merging audio and visual information in the same architecture rather than relying on post-hoc reasoning. That is, the defense model has the capability to understand the visual content when inaudible commands are present and can make inference almost simultaneously. 
Finally, the defense hypothesis is that using visual information can detect illegal inaudible commands, which would be effective in a real-world scenario as visual information can be provided by multiple-cameras for robustness and fault tolerance.


Defense techniques have already been devised against adversarial attacks \cite{tramer2018ensemble}. However, less attention has been given to developing multimodal methods. To present our proposed method, we outline audio‒visual feature fusion methods to detect inaudible voice commands. Fusion is the key research issue in multimodal, which integrates information extracted from different 
single modality data into a single feature representation \cite{lahat2015multimodal}. Then our work is inspired by the XFlow \cite{cangea2019xflow} for cross-modal deep neural networks that suggests a higher correlation between the two modalities can indicate more semantic information and higher classification accuracy. 


The attack can happen when the VCS receives the speech command, then a centre camera starts to scale the real-time road view. The command by the microphone and video (as image frames) by the camera will be sent to the multimodal defense DNN for prediction. The predictive result classifies normal speech and imperceptible speech for decision making. For example, the multimodal defense network accepts both the voice command “stop” and the traffic sign stop as inputs to generate the detection results. In this way, we are able to accurately detect illegal voice commands and 
prevent fatal accidents. The primary challenge is to design DNN via cross-modal connections for feature fusion that can effectively learn feature representations. We consider that audio and visual modalities can be correlated to benefit the defense system and protect the VCS in autonomous driving.


\subsection{Defense Network Architecture}
Our proposed detection method exploits the correlation between audio commands and visual image streams from autonomous driving recordings. The correlation is used as a condition for the acceptance of the audio commands, which makes contrary to inaudible audio attack contents. Our aim is therefore to determine if an input image and a short audio clip are semantically consistent. For simplicity, we use single frame road sign images rather than video streams as input. 

\begin{figure}[htbp]
\centering
\includegraphics[width=.5\textwidth]{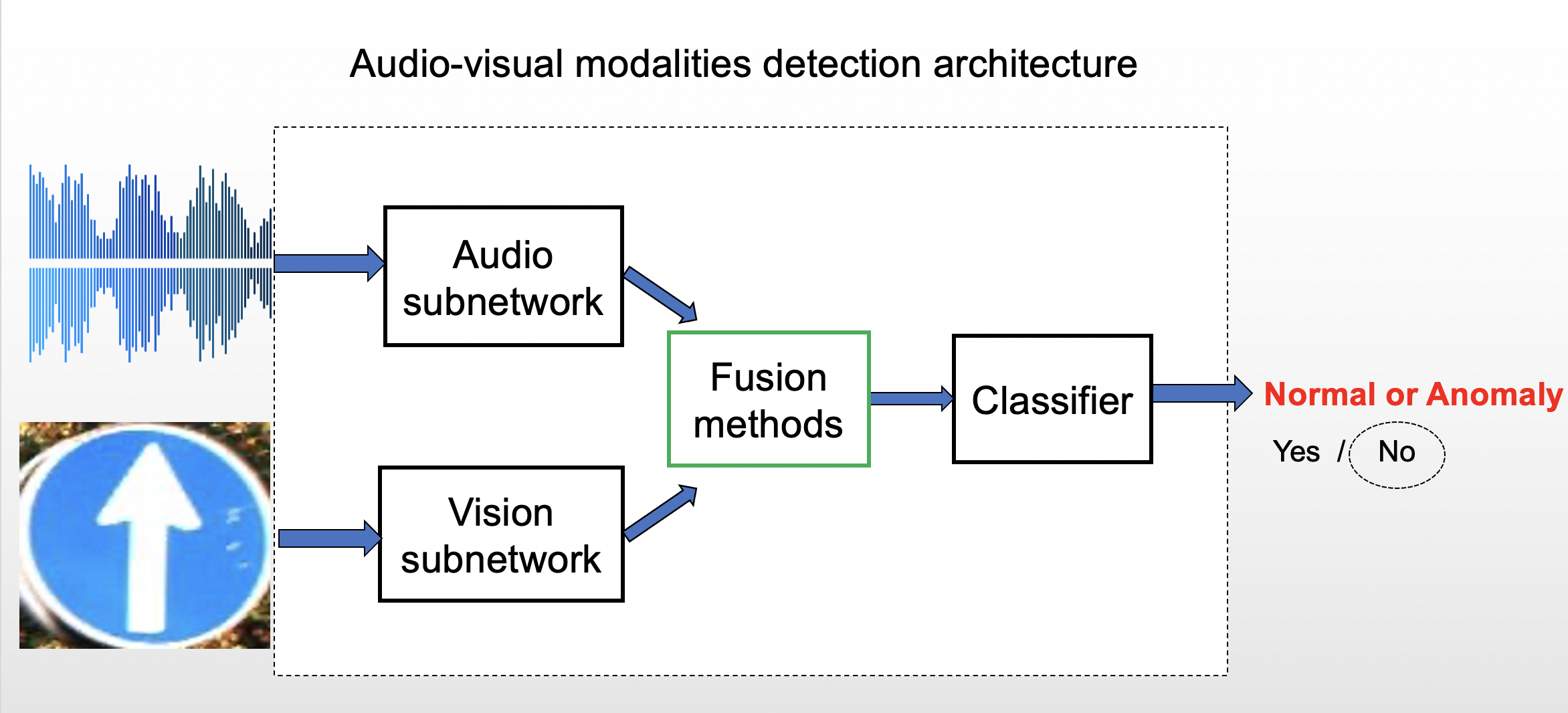}
\caption[Audio-visual detection architecture]{Audio-visual anomaly detection that learns to determine whether a pair of audio clip and image correspond to each other or otherwise.}
\label{fig 2}
\end{figure}

Fig. \ref{fig 2} shows the detection network architecture. It produces a positive prediction when an audio segment and an image are extracted from the same semantic context in a scene and a negative prediction otherwise. The framework has three distinct components: the image information, the voice commands and fusion methods: the image neural network is a Convolutional Neural Network (CNN) which is an image feature extractor; and the audio neutral network is a Multilayer Perceptrons (MLP) network. In the preprocessing phase, we extracted audio wave features into the Mel Frequency Cepstral Coefficents (MFCC) as inputs. Then the fusion layer takes audio and visual features respectively into account and generates the output.
 
\subsection{Fusion Neural Networks} 
Based on the audio-visual multimodal architecture, we create the following specific fusion networks, mainly using these technologies: attention mechanism, deconvolution and bilinear pooling, and residual fusion. Attention \cite{kim2018bilinear} in multimodal-learning provide an efficient way to utilize given visual information selectively. The Deconvolution \cite{zhang2018exfuse} is inverse operation of convolution and adapted for 1 dimension signals in multimodal sensor fusion. Bilinear pooling \cite{lin2015bilinear} is used to achieve two multimodal intermediate feature fusion. Residual fusion \cite{liu2017learning} is the learning process that improves the embedding of features while retaining the shared parameters. We have six defense networks to detect inaudible command attacks.

\begin{figure}[htbp]
\centering
\includegraphics[width=3.5in]{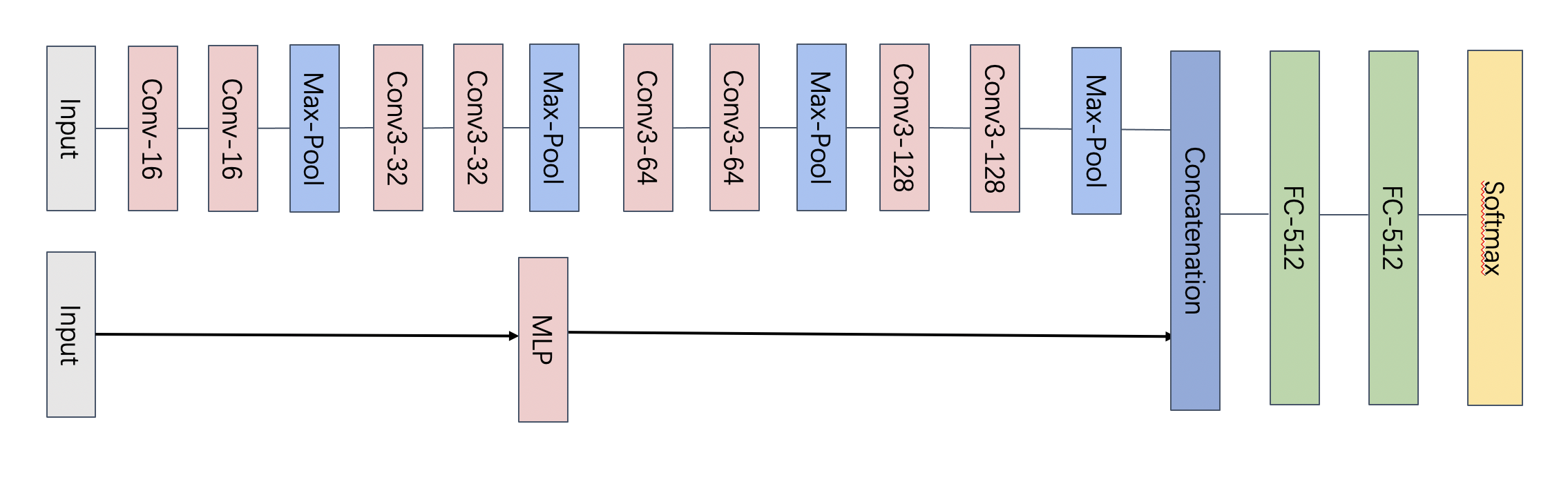}
\caption[Defense network baseline]{The baseline of the fusion network.}
\label{fig 3}
\end{figure}

1) \textbf{CNN\_MLP\_Baseline}. Convolution is usually adopted to extract features from a local neighborhood on the images. The convolutional computing adds an addictive bias and an activation function used for the feature map \cite{kong2019actionrecognition}. All convolutional and MLP layers in the proposed architecture use rectified linear unit (ReLU) \cite{li2017convergence} activations. To reduce overfitting, the batch normalization is employed after each convolutional layer and MLP layer. Then the resulting features are passed to the next processing stream max\_pooling layer and dropout \cite{srivastava2014dropout} with \emph{\textbf{P}}=0.25. Feature connections are then merged with intermediate representations corresponding to each sub-network in the target workflow. Entire features are finally classified by another MLP with a following SoftMax layer. We set this architecture as the baseline in our experiment and the architecture is shown in Fig \ref{fig 3}. 

 
\begin{figure}[htbp]
\centering
\includegraphics[width=3.5in]{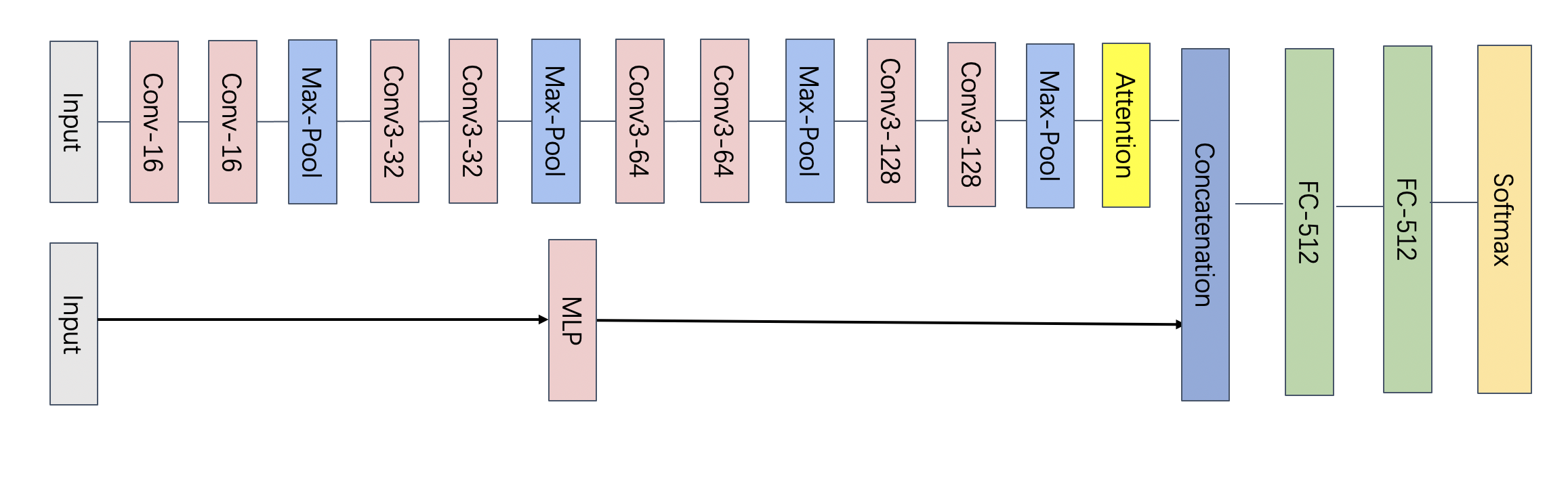}
\caption[Attention fusion defense network]{The attention mechanism fusion during two modalities.}
\label{fig 4}
\end{figure}

2) \textbf{CNN\_Attention\_MLP}. To further broaden our assumption, we extend the baseline network and use the attention mechanism for multi-modal fusion. There are two reasons: attention mechanism is the core concept of navigating focus that pays attention to certain vectors \cite{vaswani2017attention}. Also spatial and channel information as high-level features can benefit feature fusion. Hence we select Convolutional Block Attention Module (CBAM) \cite{woo2018cbam}, which is considered intermediate feature attention maps from spatial and channel dimensions. Furthermore CBAM is an effective lightweight attention module and could be seamlessly integrated into any DNN architecture. This ensures that the multi-modal network effectively learnt the image features via the attention layer. To formalize the attention mechanism for defending against inaudible command attacks, we present the architecture in Fig \ref{fig 4}. 

\begin{figure}[htbp]
\centering
\includegraphics[width=3.5in]{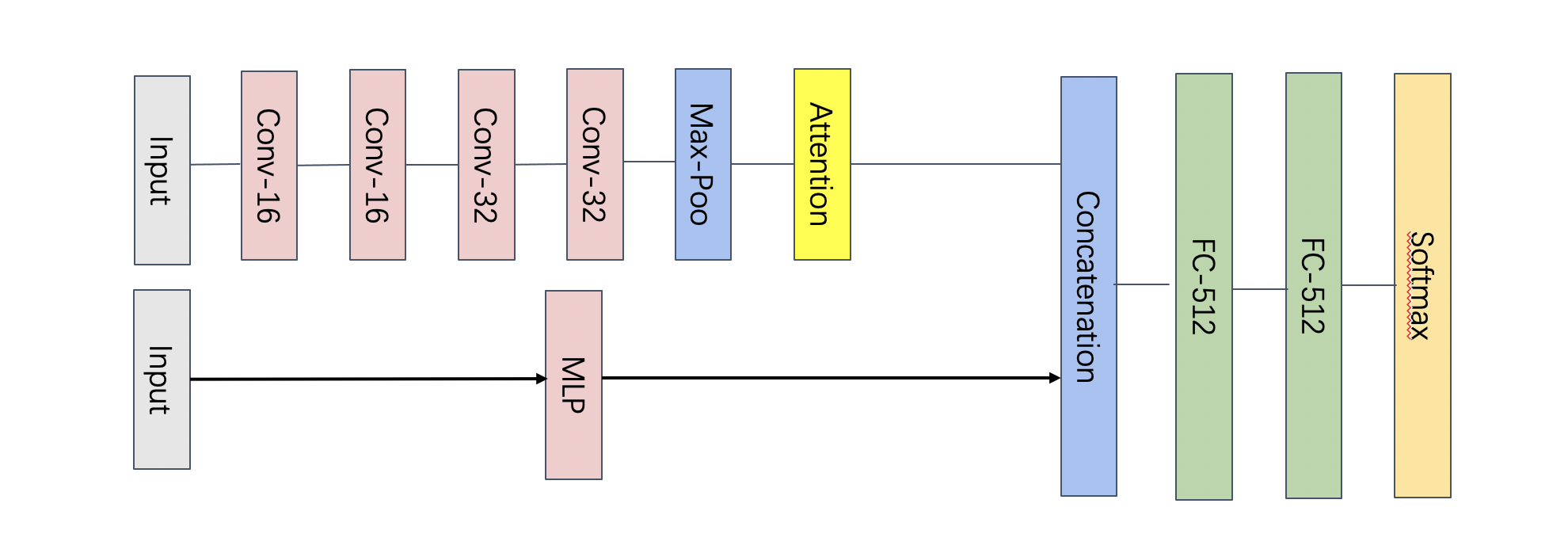}
\caption[Simplified attention fusion network]{The attention mechanism during fusion with a simplified CNN.}
\label{fig 5}
\end{figure}

3) \textbf{CNN\_BLOCK\_MLP}. Typical CNN \cite{krizhevsky2017imagenet} architecture contains a convolutional layer, pooling layer and activation layer as basic operations. The convolution operations use multiple kernels to extract features (feature maps) from the input dataset, preserving spatial information through receptive field. The pooling operation is an approach to reduce the dimensionality of feature maps followed by a convolution operation, which is used to downsample generated feature maps.
Activation operations 
introduce non-linearity into the output of neurons.
For lightweight purposes, we further propose a CNN block fusion method. It contains less convolutional layers and one attention layer that resembled the architecture used in Fig. \ref{fig 5}.

\begin{figure}[htbp]
\centering
\includegraphics[width=3.5in]{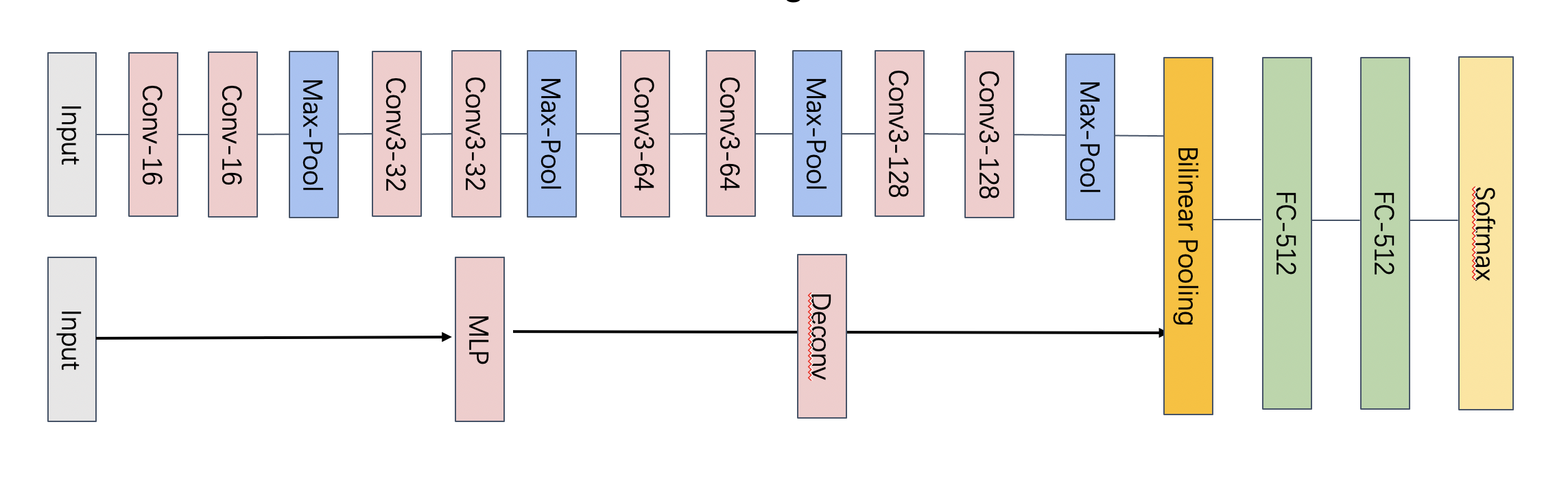}
\caption[Bilinear pooling fusion network]{The Deconvolution process for audio features and bilinear pooling for fusion.}
\label{fig 6}
\end{figure}

4) \textbf{CNN\_Deconv\_CBP\_MLP}. The motivation of this defense network is that the feature fusion network baseline is too coarse. To rectify this problem, the fully connected layer can be converted to an equivalent 2D convolutional feature representation. In terms of intermediate feature fusion design, we perform a simple deconvolutional layer for audio data. While running CNN\_Deconvolution\_MLP, the matching relations are considered as the same 2D features to explore the inter-modal correspondences between the image and audio. We use deconvolution to capture the audio feature, which is expected to be helpful for feature fusion and to ensure the structure is appropriate. In addition, Bilinear pooling is the fusion technology that merges two modalities with 2D feature pattern. Thus, we employ Deconvolution for Compact Bilinear Pooling (CBP) \cite{gao2016compact}. This defense neural network is shown in Fig \ref{fig 6}.

5) \textbf{CNN\_X\_MLP}. XFlow \cite{cangea2019xflow} consists of multimodal cross-connections deep learning architecture for feature extraction. The goal in essence attempts to exchange feature representations and exploit correlations between the audio and image datasets, for example, 1D feature to 2D feature by Deconvolution to cross-modal fusion. Meanwhile, it flattens the 2D feature to the 1D feature for cross-modal fusion. The intermediate features are connected by residual directly from raw input data. After merging intermediate representations with the corresponding modality, the output is processed by a full-connected layer. It reports that the crossmodal architecture is beneficial to improve classification performance. Thus there is a strong motivation to borrow ideas from XFlow architecture \uppercase\expandafter{\romannumeral1} as one of the defense networks to detect unauthorized voice commands.

6) \textbf{CNN\_BNN\_MLP}. So far we have used different learning representations for modalities fusion classification. However, this still requires computing resources due to each pair of input modalities that is being prohibitively expensive in autonomous driving. Hence, to reduce computational complexity, it can be helpful to use Binarized Neural Networks (BNN) \cite{hubara2016binarized}. The basic idea of BNN is to set weights and activation values as 1 or ‒1. BNN has a similar architecture to that of CNN insofar as it can be integrated into the defense network. Therefore, we employ BNN to replace the baseline neural network as the last experiment.

\section{Experiment}
\label{sec:experiment}
\subsection{Dataset}
We used the following two datasets for two modalities in our experiments: Google Speech Command version 0.01 \cite{warden2018speech} and the German Traffic Sign Recognition Benchmark (GTSRB) \cite{stallkamp2011german}. The Google Speech Command dataset is a collection of audio samples containing a set of one-second speech records of 30-words released in 2017. The dataset has a limited vocabulary and suitable for commands in controlling robotics and IoT devices. The German Traffic Sign Recognition Benchmark is a benchmark dataset for multicategory classifications. It contains more than 40 classes of traffic signs with over 50,000 images. The image size varies from 15$\times$15 to 250$\times$250 pixels. 


Each audio clip is stored as a wave-format file with a bit rate of 16Kbit/s. We used MFCC features from the spoken speech command audio clips as MFCC is commonly used to process audio signals that are in frequency and time domains, and it transforms 1D dimensional waves into a group of audio features. We set a 25-millisecond sliding window to extract audio features and a 10-milliseconds overlap between contiguous frames. We then used a Fast Fourier Transform (FFT) to generate a spectrogram and apply 128 filters for the mel-scale bank to dot product with a spectrogram. To closely emulate the acoustics effect, we applied a logarithm and Discrete Cosine Transform (DCT) so that most of the signals are located in a low-frequency range. 

To align the two datasets to mimic the two modalities in ADAS, we paired them in a procedure as follows: First, we created the audio‒visual pair as a new dataset named AID. The audio data in the Google Speech Command dataset spans four classes representing the words: ``go'', ``right'', ``left'', and ``stop''. Then we chose the related traffic sign images from the GTSRB. Each class of audio clip was associated with a related traffic sign image that contained the same content. This was to ensure that we had a variety of data that could be aligned and was suitable for machine-learning classification. As a result, the AID dataset contains the following elements:
\begin{enumerate}
    \item the image data corresponding to the 2D single frame (There are 1,200 traffic signs with 300 images for each class);
    \item the audio data represented by MFCCs.
\end{enumerate}

\begin{figure}[!t]
\begin{minipage}[c]{\linewidth}
\subfloat[]{\includegraphics[width=.45\textwidth]{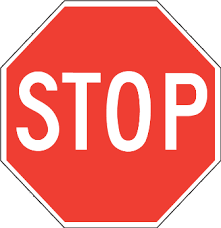}\label{fig 7a}}\hfil
\subfloat[]{\includegraphics[width=.45\textwidth]{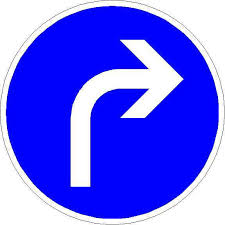}\label{fig:7b}}\hfil 
\end{minipage}
\vspace{-5pt}
\caption{Examples of road signs used}
\label{fig 7}
\end{figure}

We consider a mismatched traffic sign and corresponding audio input as anomaly. The dataset for anomalies was created from random samples of mismatched audio and image data. Fig. \ref{fig 7} (a) and (b) present the ``stop" audio wave with ``stop" road sign and unrelated ``right" road sign. 



\subsection{Results}
As the raw image data had different widths and heights, we resized them to a uniform size of 64×64 pixels. For audio feature processing, since the fully connected layer has a pre-defined dimension for input data, the size of the input audio data must have a fixed size in order to match dimensions. We tried to use the first 1000 dimensions as audio input. Our implementation used Keras with Tensorflow as the backend. We split our data into a training and a validation set. In the training process, we set the batch size to 128 and used Adam optimizer on the training data for 300 epochs with a cross-entropy loss function. We tested these models on each epoch and reported the best accuracy in the end. Two 2080 Ti Nvidia GPU with 16 G memory were used for the experiments. All the multi-modal experiments employed cross-validation to overcome overfitting. We have evaluated the six proposed 
audio-visual information fusion network architectures in our experiments. 

We trained these models on the dataset containing one class of normal audio/image pairs and another with mismatched audio/image pairs. We then evaluated these models on two validation datasets. The first one contained normal pairs and the second one contained mismatched pairs. The mismatched pairs represented the scenarios where the ADAS was under inaudible voice command attack. The evaluation results are shown in Table \uppercase\expandafter{\romannumeral1} for the normal cases and in Table \uppercase\expandafter{\romannumeral2} for attacking cases.

To measure the performance of our proposed defense networks, we used \emph{Accuracy}, \emph{Precision}, \emph{Recall} and \emph{F1}. In the evaluation, we used weighted metrics from scikit-learn for the Precision and Recall. Our proposed defense networks achieved the high accuracy, precision and recall for classifying normal cases as shown in Table \uppercase\expandafter{\romannumeral1}. Table \uppercase\expandafter{\romannumeral2} had shown these results under inaudible attacks. The CNN\_Attention\_MLP was slightly higher than the other networks. Under inaudible attacks, the defense system was able to achieve an accuracy above 89\% in identifying attacks. Even though CNN\_MLP\_Baseline had the highest F1 score, it was worth noting that CNN\_Attention\_MLP was a lightweight network and potentially more suitable for running on edge devices. Although BNN had the smallest model size, its accuracy was significantly lower than other models. In general, our method was effective in detecting inaudible voice command attacks while incurring low error rates in misclassifying normal audio/image pairs. 

\begin{table}[htbp] 
\scriptsize
\begin{center} 
\begin{tabular}{|l|l|l|l|l|p{10cm}}  
\hline  
Network & Accuracy & Precision & Recall & F1-measure \\ \hline  
CNN\_MLP\_Baseline &  98.09\% &  98.12\% &  98.06\% & 98.09 \% \\ \hline 
CNN\_Attention\_MLP &  99.03\% & 99.09\% &  98.98 \% & 99.03\%\\  \hline 
CNN\_Deconv\_CBP\_MLP  & 98.96\%  & 99.03\% & 98.91\% & 98.97\% \\ \hline 
CNN\_X\_MLP  &  98.80\% &  98.88\% &  98.76\% &  98.82\% \\ \hline 
CNN\_BLOCK\_attention\_MLP & 98.76\% &  98.79\% &  98.67\% & 98.73\% \\ \hline 
\end{tabular}  
\end{center}  
\caption{Accuracy, Precision, Recall and F1-measure without attacks}  
\end{table}

\begin{table}[!t] 
\scriptsize
\begin{center}  
\begin{tabular}{|l|l|l|l|l|p{10cm}}  
\hline   
Network & Accuracy &  Precision & Recall & F1-measure \\ \hline  
CNN\_MLP\_Baseline & 89.2\% & 91.3\% &  89.2\% &   89.2 \% \\ \hline 
CNN\_Attention\_MLP & 88.9\% &  91.1\% & 88.9\% &  88.9 \% \\ \hline 
CNN\_Deconv\_CBP\_MLP  & 88.5\% & 90.8\%  & 88.5 \% & 88.4\%  \\ \hline 
CNN\_X\_MLP & 88.9\% & 90.9\% &  88.9\% &  88.7\% \\ \hline 
CNN\_BLOCK\_attention\_MLP & 89.2\% & 91.2\% &  89.2\% &  89.1\% \\ \hline 
CNN\_BNN\_MLP & 68.2\%  & 77.8\%  &  68.2\% &  65.8\% \\ \hline 
\end{tabular}  
\end{center}  
\caption{Precision, Recall and F1-measure under attacks}  
\end{table}

\begin{figure}[htbp]
\centering
\includegraphics[scale=1.0]{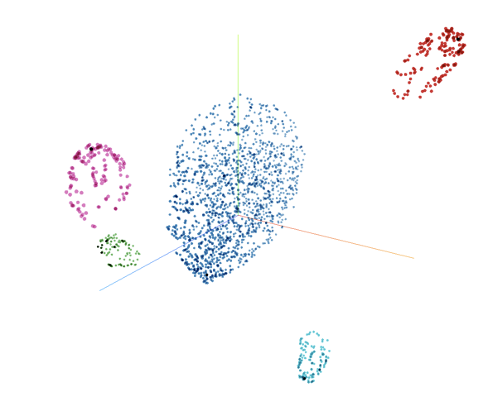}
\caption[Clustering results]{Deep blue clustering is the inaudible command with incorrect visual information, and the rest are normal audio-visual clusterings.}
\label{figl 9}
\end{figure}

To show the performance of CNN\_Attention\_MLP. We used t-Distributed Stochastic Neighbor Embedding (t-SNE) \cite{van2008visualizing} for 3D visualizing. The clustering results on t-SNE output are shown in Fig. \ref{figl 9}. The margins between clusters further show inaudible attacks can be singled out by our approach.

\section{Discussion}
\label{sec:discussion}
Inference using deep learning models on edge devices in an ADAS often incurs expensive computational cost. Overcoming this challenge requires finding a balance between accuracy and inference speed. In recent years, great progress has been made to run deep learning models on IoT devices efficiently. These lightweight architectures are deployed in IoT systems and are considered to be important to preserve runtime integrity. Various lightweight architectures are proposed for this setting. Our proposed methods intend to simplify model architecture with just sufficient number of model parameters to achieve satisfactory inference accuracy. With the advance in techniques of edge devices capable of running certain deep learning models \cite{lee2021opportunistic}, our proposed inaudible defense system can be deployed on edge devices in real world.

Our fusion defense method uses camera data to identify suspicious voice commands. However, the classification of camera data can be wrong on its own. This can happen either because the camera inputs can be misclassified by the model (e.g., not seen before) or there are adversarial attacks on the camera inputs.
Such scenarios would trigger false alarms. Frequent false alarms may cause problems for drivers. Reducing false alarms by utilizing multiple camera inputs and more robust models is our future work.

\section{Conclusion}
\label{sec:conclusion}
In this paper, we consider inaudible voice commands as a new threat for ADAS. Attackers can use ultrasounds to steal private information. To solve this problem, we are the first to propose a totally new insight of using multi-modal models as the defense strategy. Our design of lightweight algorithms and experiments demonstrate that empirically defense networks can learn cross-modal knowledge and have powerful ability of visual reasoning for detecting voice attacks. 

We believe that multi-modal sensor fusion is a frontier for future research and hope our research will extend voice adversarial defense analysis to improve model robustness and build reliable techniques in ADAS. Given that autonomous driving is a rapidly developing technology to be widely adopted, our work can contribute to improve the state-of-the-art defence strategy against often stealthy autonomous driving attacks.
Future work
includes using multiple camera images as inputs to create more realistic multi-modality sensor fusion models. We will also focus on finer-granularity vision objects detection (e.g., lane detection) to detect more subtle inaudible adversarial attacks. 


\section{ACKNOWLEDGE}
This work is in part supported by a CSIRO Data61 Collaborative Research Project (CRP) C020996 and an Australian Research Council Project (ARC) LP190100676.

\bibliographystyle{IEEEtran}

\end{document}